\documentclass[]{interact}

\usepackage{epstopdf}
\usepackage[caption=false]{subfig}

\usepackage[numbers,sort&compress]{natbib}
\bibpunct[, ]{[}{]}{,}{n}{,}{,}
\renewcommand\bibfont{\fontsize{10}{12}\selectfont}
\makeatletter
\def\NAT@def@citea{\def\@citea{\NAT@separator}}
\makeatother

\theoremstyle{plain} 

\theoremstyle{definition}

\theoremstyle{remark}

\usepackage{graphicx}
\usepackage{hyperref}
\usepackage{multirow}
\usepackage{booktabs}
\usepackage{array}

\begin{document}

\articletype{RESEARCH PAPER}

\title{Semi-automatic staging area for high-quality structured data extraction from scientific literature}

\author{
    \name{Luca Foppiano\textsuperscript{a,b}\thanks{Corresponding authors: Luca Foppiano (luca@foppiano.org) and Masashi Ishii (ISHII.Masashi@nims.go.jp)}, Tomoya Mato\textsuperscript{a}, Kensei Terashima\textsuperscript{c}, Pedro Ortiz Suarez\textsuperscript{d}, Taku Tou\textsuperscript{c}, Chikako Sakai\textsuperscript{c}, Wei-Sheng Wang\textsuperscript{c}, Toshiyuki Amagasa\textsuperscript{b}, Yoshihiko Takano\textsuperscript{c}, Masashi Ishii\textsuperscript{a}}
    \affil{\textsuperscript{a}Materials Modelling Group, Data-driven Materials Research Field, Centre for Basic Research on Materials, NIMS, JP; 
    \textsuperscript{b}Knowledge and Data Engineering, Centre for Computational Sciences, University of Tsukuba, JP;
    \textsuperscript{c}Frontier Superconducting Materials Group, MANA, NIMS, Tsukuba, JP; \textsuperscript{d}DFKI GmbH, DE}
}

\maketitle

\begin{abstract}
We propose a semi-automatic staging area for efficiently building an accurate database of experimental physical properties of superconductors from literature, called SuperCon\textsuperscript{2}, to enrich the existing manually-built superconductor database SuperCon. 
Here we report our curation interface (SuperCon\textsuperscript{2} Interface) and a workflow managing the state transitions of each examined record, to validate the dataset of superconductors from PDF documents collected using Grobid-superconductors in a previous work~\cite{lfoppiano2023automatic}. 
This curation workflow allows both automatic and manual operations, the former contains ``anomaly detection'' that scans new data identifying outliers, and a ``training data collector'' mechanism that collects training data examples based on manual corrections. 
Such training data collection policy is effective in improving the machine-learning models with a reduced number of examples. 
For manual operations, the interface (SuperCon\textsuperscript{2} interface) is developed to increase efficiency during manual correction by providing a smart interface and an enhanced PDF document viewer. 
We show that our interface significantly improves the curation quality by boosting precision and recall as compared with the traditional ``manual correction''.  
Our semi-automatic approach would provide a solution for achieving a reliable database with text-data mining of scientific documents.

\end{abstract}

\begin{keywords}
    materials informatics, superconductors, machine learning, database, tdm
\end{keywords}

\section{Introduction}
The emergence of new methodologies using machine learning for materials exploration has given rise to a growing research area called materials informatics (MI)~\cite{10.3389/fchem.2022.930369}.
This field leverages the knowledge of the materials data accumulated in the past to efficiently screen candidates of the materials with desired properties.
As a matter of course, such an approach requires a larger amount of material-related data for training models.
Researchers have been developing large aggregated databases of physical properties generated by first-principles calculations based on Density Functional Theory (DFT), such as Materials Project~\cite{materialsprojectJain2013}, JARVIS (Joint Automated Repository for Various Integrated Simulations)~\cite{aflowcurtarolo2012aflow}, NOMAD (Novel Materials Discovery)~\cite{nomad}, that played a role of a strong driving force for the development of materials informatics. 
Using DFT data for machine learning (ML) in materials science has become popular since, in principle, it allows researchers to simulate and obtain various types of physical properties of the target materials only by knowing the crystal structures of the subjects. 
Those DFT codes are designed to reproduce/simulate the physical properties that should be observed by experiments in reality.
Nonetheless, caution must be exercised while utilising these computed figures for constructing ML models aimed at steering experiments. 
This caution arises due to the potential lack of validity in their predictions when dealing with specific simplifications of the interactions between atoms and electrons in solids, such as electron-electron Coulomb correlation, spin-orbit coupling, and similar factors.

On the contrary, accumulated datasets of experimental data from scientific publications are still scarce, despite abundant publication availability, and exponential growth in materials science~\cite{Pratheepan_2019}.
Currently, only a few limited resources exist, such as the Pauling File~\cite{Blokhin2018ThePF_paulingFile} and SuperCon~\cite{ishii2023structuring}, necessitating reliance on manual extraction methods. 
This scarcity can be attributed to inadequate infrastructure and a shortage of expertise in computer science within the materials science field.


The SuperCon database was built manually from 1987~\cite{ishii2023structuring} by the National Institute for Materials Science (NIMS) in Japan and it is considered a reliable source of experimental data on superconductors~\cite{roter2020predicting, stanev_machine_2017, tran2022machine, konno2021deep}. 
However, the updates of SuperCon have become increasingly challenging due to the high publication rate. 
In response to the need for a more efficient approach to sustain productivity, we embarked on the development of an automated system for extracting material and property information from the text contained in relevant scientific publications. 
This automated process enabled the rapid creation of ``SuperCon\textsuperscript{2} Database'', a comprehensive database of superconductors containing around 40000 entries, within an operational duration of just a few days~\cite{lfoppiano2023automatic}. 
Matching the level of quality seen in SuperCon while simultaneously automating the extraction of organised data can be achieved with a properly designed curation process. 
We use the term \emph{curation} to describe the overall process of reviewing and validating database records, while \emph{correction} refers to the specific action of altering the values of one or more properties within an individual record.
At the moment of writing this article, we are not aware of any other curation tool focusing on structured databases of extracted information. 
There are several tools for data annotation, such as Inception~\cite{klie-etal-2018-inception}, and Doccano~\cite{doccano} which concentrate on text labelling and classification.

In this work, we designed and developed a workflow with a user interface, ``SuperCon\textsuperscript{2} Interface'', crafted to produce structured data of superior quality and efficiency to the one obtained by the ``traditional'' manual approach consisting of reading documents and noting records, usually on an Excel file.
We developed this framework around the specific use case of SuperCon, however, our goal is to be adapted to alternative data frameworks.

Our contributions can be summarised as follows:
\begin{itemize}
    \item We developed a workflow and a user interface that allow the curation of a machine-collected database. We demonstrate that using it for data correction resulted in higher quality than the ``traditional'' (manual) approach.
    \item We devise an anomaly detection process for incoming data lower rejection rate (false positive rate) from domain experts.
    \item We propose a mechanism that selects training data based on corrected records, and we demonstrate that such selections are rapidly improving the ML models.
\end{itemize}

The subsequent sections, Section~\ref{sec:curation-workflow} describes the curation workflow and Section~\ref{sec:user-interface} the user interface on top of it.
Finally, we discuss our evaluation experiments and results in Section~\ref{sec:results-and-evaluation}.  

\section{Curation workflow}
\label{sec:curation-workflow}
\begin{figure}[ht]
  \centering
  \includegraphics[width=1\textwidth]{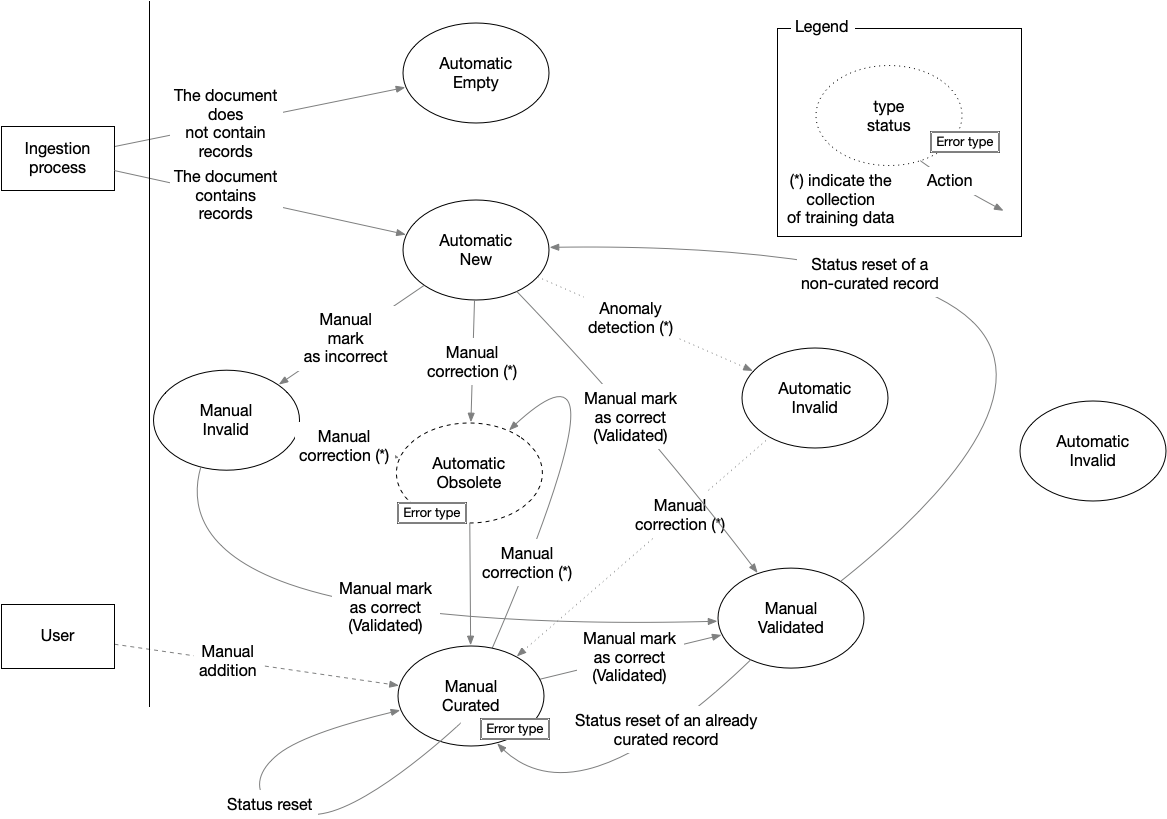} 
  \caption{Schema of the curation workflow. Each node has two properties: type and status (Section~\ref{subsec:curation-status}). Each edge indicates one action. The workflow starts on the left side of the figure. The new records begin with ``Automatic, New''. Changes of state are triggered by automatic (Section~\ref{subsec:anomaly-detection}) or manual operations (update, mark as valid, etc.. Section~\ref{subsec:manual_correction}) and results in changes of the properties in the node. Each combination of property values identifies each state. ``(*)'' indicates a transition for which the training data are collected (Section~\ref{subsec:feedback-loop-training-data})}
  \label{fig:curation-workflow}
\end{figure}

The curation of the SuperCon\textsuperscript{2} Database acts as a workflow where user actions result in database records state transitions (Figure~\ref{fig:curation-workflow}). 
Allowed manual actions include a) \textit{mark as valid} (validation) when a record is considered correct or corrected by someone else. When a record is not valid, users can: b) \textit{mark as invalid} when considered ``potentially'' invalid (or the curator is not confident), c) perform \textit{manual correction} to update it according to the information from the original PDF document, and d) \textit{remove} the record when it was not supposed to be extracted.

Besides manual operations from users, this workflow supports also automatic actions: ``anomaly detection'' for pre-screening records (Section~\ref{subsec:anomaly-detection}) and the ``training data collector'' for accumulating training data for improving ML models (Section~\ref{subsec:feedback-loop-training-data}). 


Although only the most recent version of a record can be viewed on this system, the correction history is recorded (Section~\ref{subsec:curation-and-processing-logs}). 



\subsection{Workflow control}
\label{subsec:workflow-control}
The workflow state is determined by the ``curation status'' (Section~\ref{subsec:curation-status}), the user action, and the error type (Section~\ref{subsec:error-types}).

\subsubsection{Curation status} 
\label{subsec:curation-status}
The curation status (Figure~\ref{fig:curation-workflow}) is defined by \emph{type} of action, manual or automatic, and \emph{status}, which can assume the following values: 
\begin{itemize}
    \item \textbf{new}: default status when a new record is created.
    \item \textbf{curated}: the record has been amended manually.
    \item \textbf{validated}: the record was manually marked as valid.
    \item \textbf{invalid}: the record is wrong or inappropriate for the situation (e.g., T\textsubscript{m} or T\textsubscript{curie} extracted as superconducting critical temperature).
    \item \textbf{obsolete}: the record has been updated and the updated values are stored in a new record (internal status\footnote{``internal status'' indicates that their records should be hidden in the interface}).
    \item \textbf{removed}: the record has been removed by a curator (internal status).
\end{itemize}


\subsubsection{Error types}
\label{subsec:error-types}
We first introduced \emph{error type} in~\cite{lfoppiano2023automatic} and extended their scope in this work to consider data curation and anomaly detection. 

Users are required to select one \emph{Error Type} at every record update or removal. This information is stored in the ``original'' record and can be different at every record modification.
The error type values can be summarised as follows: 

\begin{itemize}
    \item \textbf{From table}: the entities Material $\rightarrow$ T\textsubscript{c} $\rightarrow$ Pressure are identified in a table. At the moment, table extraction is not performed
    \item \textbf{Extraction}: The material, temperature, and pressure are not extracted (no box) or extracted incorrectly. 
    \item \textbf{Linking}: The material is incorrectly linked to the T\textsubscript{c} given that the entities are correctly recognised.
    \item \textbf{T\textsubscript{c} classification}: The temperature is not correctly classified as ``superconductors critical temperature'' (e.g., Curie temperature, Magnetic temperature...).
    \item \textbf{Composition resolution}: The exact composition cannot be resolved (e.g., the stoichiometric values cannot be resolved).
    \item \textbf{Value resolution}: The extracted formula contains variables that cannot be resolved, even after having read the paper. This includes when data is from tables
    \item \textbf{Anomaly detection}: The data has been modified by anomaly detection, which facilitates their retrieval from the interface.
    \item \textbf{Curation amends}: The curator is updating the data which does not present issues due to the automatic system.
\end{itemize}

\subsection{Anomaly detection}
\label{subsec:anomaly-detection}
Anomaly detection is the process of identifying unusual events or patterns in data. 
In our context, this means identifying data that are greatly different from the expected values.
This post-process was introduced in a limited scope to draw attention to certain cases during the curation.

The anomaly detection uses a rule-based approach and marks any record that matches the following conditions
\begin{itemize}
    \item the extracted T\textsubscript{c} is greater than room temperature (273 K), negative, or contains invalid characters and cannot be parsed (e.g. ``41]'')
    \item the chemical formula cannot be processed by an ensemble composition parser that combines Pymatgen~\cite{Ong2013}, and text2chem~\cite{kononova_text-mined_2019} 
    \item the extracted applied pressure cannot be parsed or falls outside the range 0 - 250 GPa.
\end{itemize}

Records identified as anomalies have \emph{status} ``invalid'' and \emph{error type} ``anomaly detection'' for easy identification.
Since this process may find false positives, its output requires validation from curators. 
For example, in certain contexts, T\textsubscript{c} values above room temperature or applied pressure up to 500 GPa may be valid in researchers' hypotheses, calculations, or simulated predictions. 

We ran the anomaly detection on the full SuperCon\textsuperscript{2} Database (40324 records~\cite{lfoppiano2023automatic}). 
The anomaly detection identified 1506 records with invalid T\textsubscript{c}, 5021 records with an incomplete chemical formula, 304 records with invalid applied pressure, and 1440 materials linked to multiple T\textsubscript{c} values. 
Further analysis and cross-references with contrasting information may be added in future. 

\subsection{Automatic training data collector}
\label{subsec:feedback-loop-training-data}
The curation process is a valuable endeavour demanding significant knowledge and human effort. 
To maximise the use of this time for collecting as much information as possible.
We integrated an automatic procedure in the curation process that, for every correction, accumulates the related data examples that can be used to improve the underlying ML models. 

\subsubsection{Training data collection}
In the event of a correction (update, removal) in a database record, this process retrieves the corresponding raw data: the text passage, the recognised entities (spans), and the layout tokens information. 
This information is sufficient to be exported as training examples, which can be examined and corrected, and feedback to the ML model. 

\subsubsection{Training data management}
\begin{figure}[ht]
  \centering
  \includegraphics[width=1\textwidth]{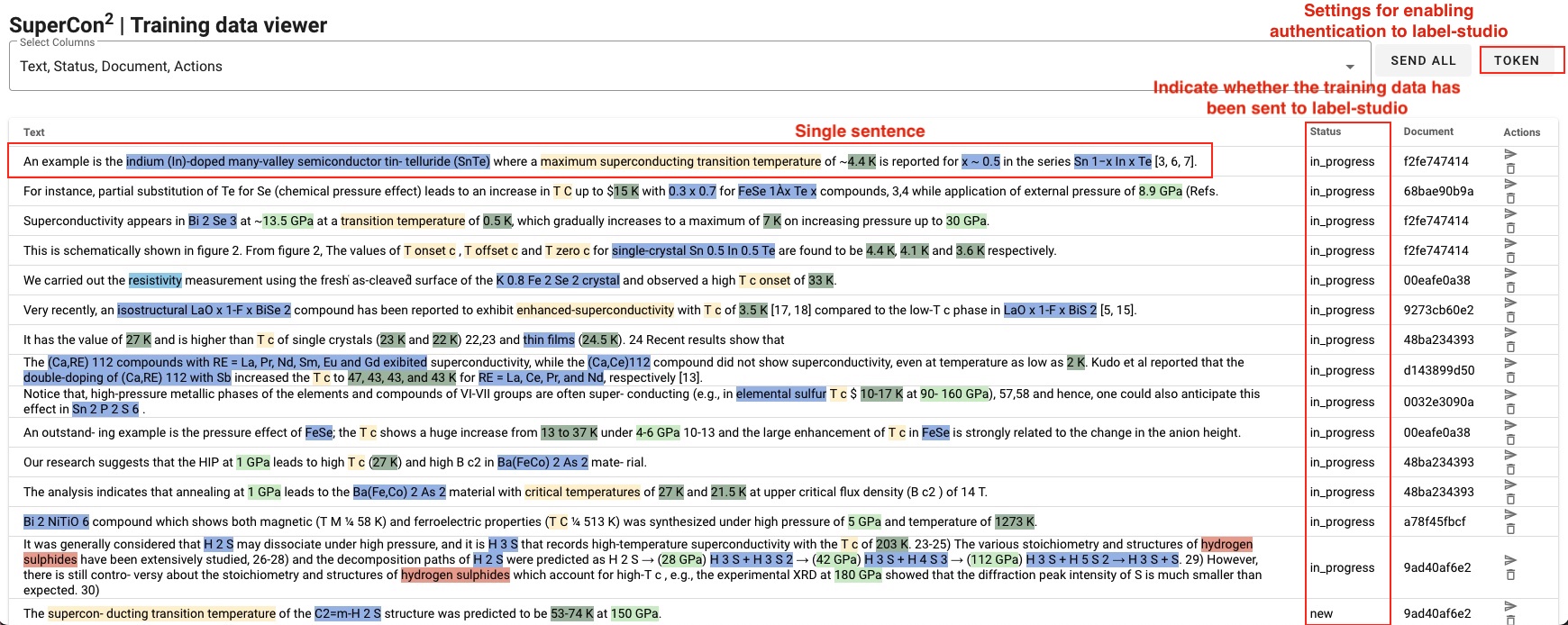} 
  \caption{Screenshot of the training data management page in the SuperCon\textsuperscript{2} interface. Each row contains one potential training data example. Each example is composed of a sentence and its extracted entities (highlighted in colour) with potential annotation mistakes that need to be corrected using an external tool: we used Label-Studio~\cite{Label_Studio}. The column ``Status'' indicate whether the example has been sent or not to the external tool.}
  \label{fig:training-data-view}
\end{figure}

We designed a specific page of the interface (Section~\ref{sec:user-interface}) to manage the collected data (Figure~\ref{fig:training-data-view}) in which each row corresponds to a training example composed by the decorated text showing the identified entities, the document identifier, and the status. 
The users can examine the data, delete it, send it to the annotation tool to be corrected, and then export them.
We integrated our interface with Label-studio~\cite{Label_Studio} for the correction of the collected training examples. 
Label-studio is an open-source, python-based, and modern interface supporting many different TDM tasks (NER, topic modelling, image recognition, etc.). 

\section{Curation interface}
\label{sec:user-interface}

\begin{figure}[ht]
  \centering
  \includegraphics[width=1\textwidth]{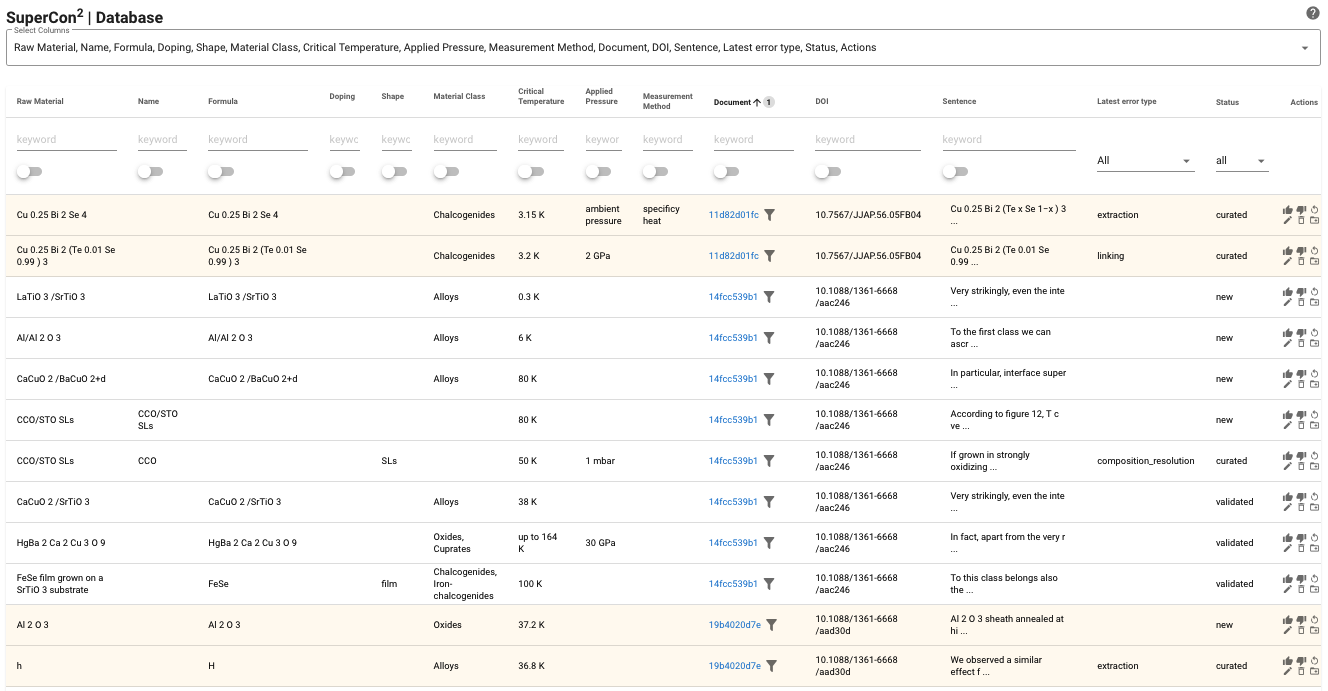} 
  \caption{Screenshot of SuperCon\textsuperscript{2} interface showing the database. Each row corresponds to one material-T\textsubscript{c} pair. On top, there are searches by attribute, sorting and other filtering operations. On the right (last column) there are curation controls (mark as valid, update, etc.).   Records are grouped by document with alternating light yellow and white. }
  \label{fig:curation-interface-database}
\end{figure}

The workflow is operated through the user interface, which offers several key features to facilitate the data curation process (Figure~\ref{fig:curation-workflow}).
It provides a comprehensive view of materials and their related properties as a table which includes search, filtering, and sorting functionality (Figure~\ref{fig:curation-interface-database}). 
The detailed schema, including examples, is reported in our previous work~\cite{lfoppiano2023automatic}.

\begin{figure}[ht]
  \centering
  \includegraphics[width=1\textwidth]{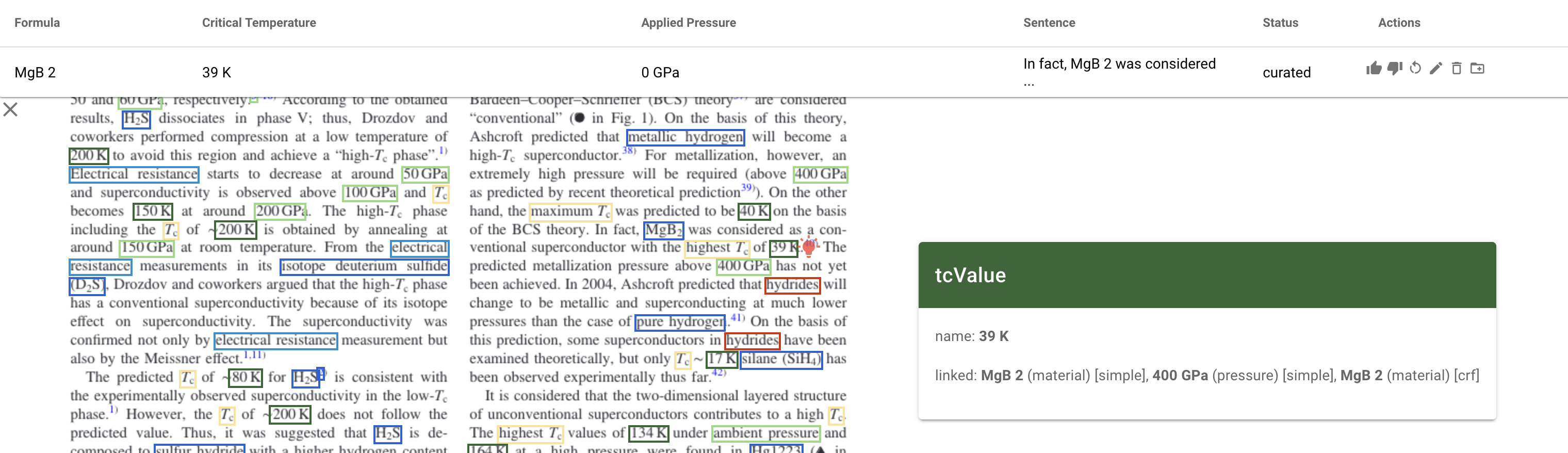} 
  \caption{PDF document viewer showing an annotated document. The table on top is linked through the annotated entities. The user can navigate from the record to the exact point in the PDF, with a pointer (the red bulb light) identifying the context of the entities being examined. }
  \label{fig:pdf-view}
\end{figure}

During the curation process, it is often necessary to switch back and forth between the database record and the related context in the paper (the related paragraph or sentence). 
Our interface provides a viewer for individual documents, which visualises in the same window a table with the extracted records and the original PDF document decorated with annotations that identify the extracted materials and properties (Figure~\ref{fig:pdf-view}). 




\subsection{Manual curation approach}
\label{sec:data-correction}
\label{subsec:manual_correction}

In this section, we discuss our strategy concerning manual curation, which is still indispensable for developing high-quality structures. 

We selected curators from domain experts in the field, to certify sufficient data quality. 
Nevertheless, as confirmed from our experiment in Section~\ref{sec:interface-evaluation}, the experience of each individual may have an impact on the final result.
We followed two principles to guarantee robustness in the curation process. 
First, we built solid curation documentation as a form of example-driven guidelines with an iterative approach we first introduced in \cite{foppiano2021supermat}. 
Then, we used a double-round validation approach, in which the data was initially corrected by one person, and validated in a second round, by a different individual. 



\subsection{Curation guidelines}
\label{subsec:curation-guidelines}

\begin{figure}[ht]
  \centering
  \includegraphics[width=1\textwidth]{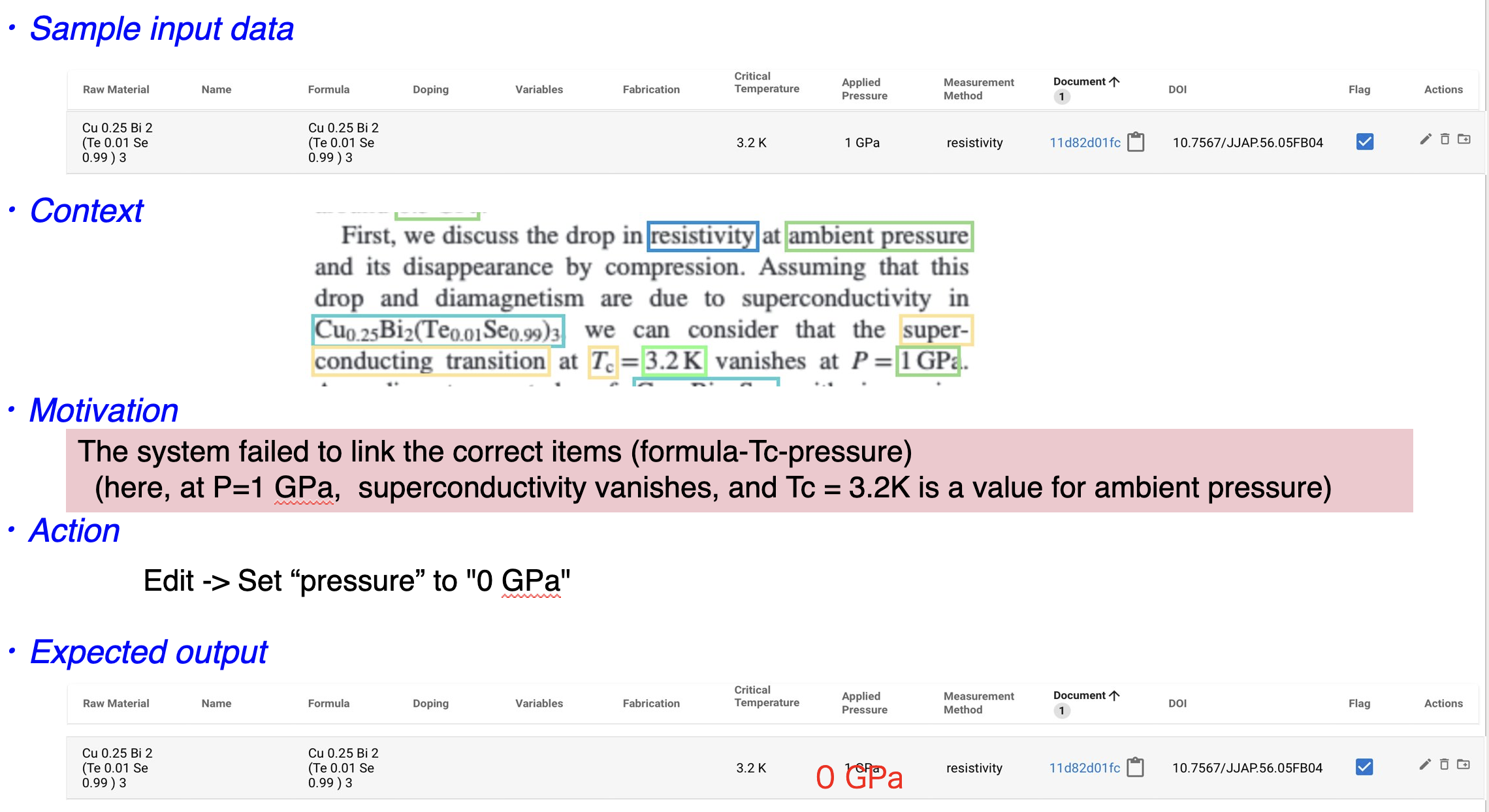} 
  \caption{Sample curation sheet from the curation guidelines. The sheet is composed of the following information: a) {Sample input data}: a screenshot of the record from the ``SuperCon\textsuperscript{2} Interface'', b) \textit{Context} represented by the related part of the annotated document referring to the record in exams. c) The \textit{Motivation}, describing the issue, d) the \textit{Action} to be taken, and the \textit{Expected output}.
 }
  \label{fig:example-curation-sheet}
\end{figure}

The guidelines consist mainly of two parts: the general principles and the correction rules with examples of solutions.
The guidelines are designed to provide general information applied to corrections and very basic explanations containing illustrations for a faster understanding (e.g. the meaning of the colours of the annotations). 
Differently from our previous work~\cite{foppiano2021supermat}, these guidelines are divided into examples for different scenarios based on the error types mentioned in Section~\ref{subsec:error-types}.
Each example described the initial record, its context, the expected corrected record and a brief explanation, as illustrated in Figure~\ref{fig:example-curation-sheet}. 

\subsection{Curation and processing logs}
\label{subsec:curation-and-processing-logs}

\begin{figure}[ht]
  \centering
  \includegraphics[width=1\textwidth]{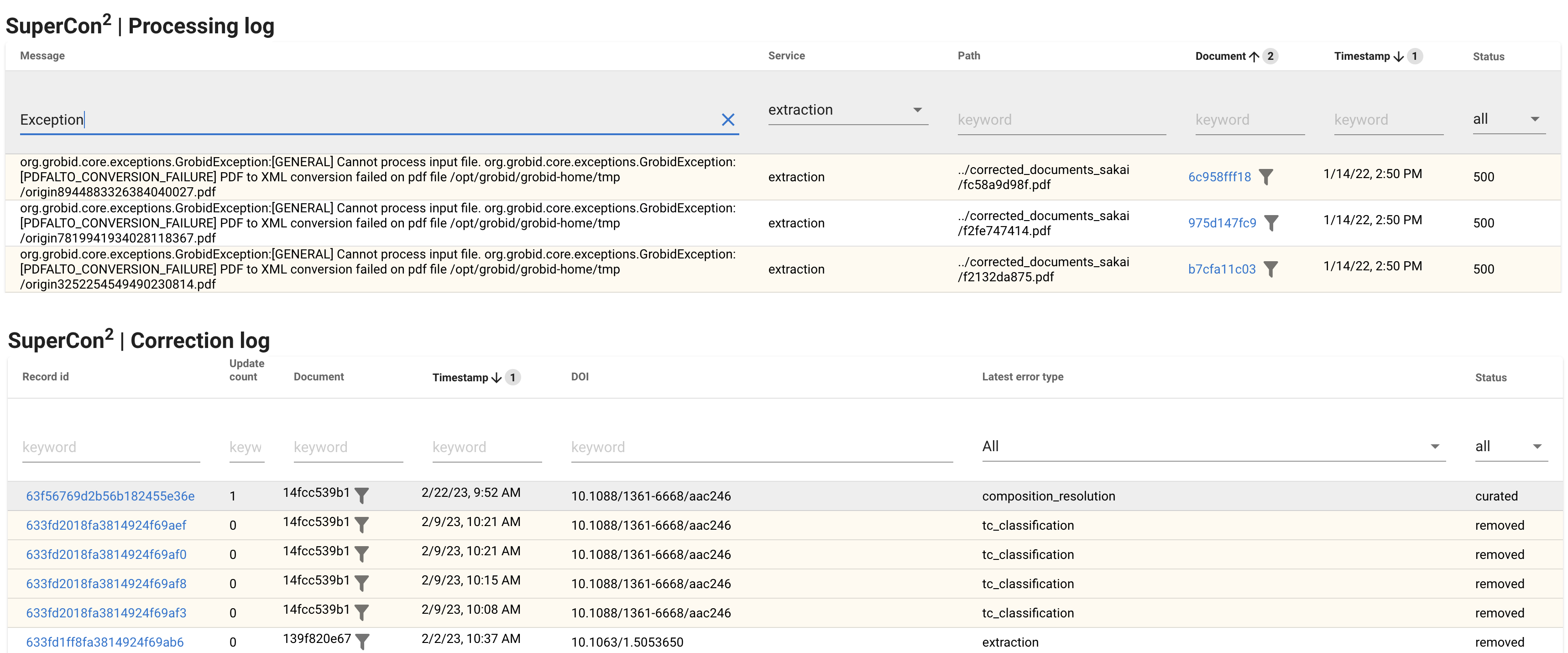} 
  \caption{Top: \textit{Processing log}, showing the output of each ingestion operation and the outcome with the detailed error that may have occurred. Bottom: \textit{Correction log}, indicating each record, the number of updates, and the date/time of the last updates. By clicking on the ``Record id'', is possible to visualise the latest record values.}
  \label{fig:processing-curation-log}
\end{figure}

The Supercon\textsuperscript{2} interface gives access to information regarding the ingestion (processing log) and the curation process (curation log). 
The processing log is filled up when the new data is ingested, it was built to have minimal functions able to explain why certain documents haven't been processed (Figure~\ref{fig:processing-curation-log} top). 
For example, sometimes documents fail because they don't contain any text (image PDF documents) or they are too big (more than 100 pages). 

The curation log provides a view of what, when and how a record has been corrected (Figure~\ref{fig:processing-curation-log} bottom).


\section{Results and evaluation}
\label{sec:results-and-evaluation}

In this section, we illustrate the experiments we have run to evaluate our work. 
The evaluation is composed of three sets of results. 
The anomaly detection rejection rate (Section~\ref{subsec:anomaly-detection-evaluation}) indicates how many anomalies were rejected by curators after validation. 
Then, we demonstrate that the training data automatically selected contributed to improving the ML model with a small set of examples (Section~\ref{subsec:training-data-generation-evaluation}) 
Finally, we evaluated the quality of the data extraction using the interface (and the semi-automatic TDM process) against the classical method of reading the PDF articles and noting the experimental information in an Excel file. In Section~\ref{sec:interface-evaluation} we find out that using the interface improves the quality of the curated data by reducing missing experimental data.

\subsection{Anomaly detection rejection rate}
\label{subsec:anomaly-detection-evaluation}

We evaluated the anomaly detection by observing the ``rejection rate'' which consists of the number of detected anomalies that were rejected by human validation. 
Running the anomaly detection on a database subset with 667 records, it found 17 anomalies in T\textsubscript{c}, 1 anomaly in applied pressure, and 16 anomalies in the chemical formulas. 
Curators examined each reported record and rejected 4 (23\%) anomalies in T\textsubscript{c}, 6 anomalies (37\%) in chemical formulas and 0 anomalies in applied pressure. 
This indicates an appropriate low rate of false positives although a study with a larger dataset might be necessary. 

\subsection{Training data generation}
\label{subsec:training-data-generation-evaluation}
We selected around 400 records in the Supercon\textsuperscript{2} Database that were marked as invalid by the anomaly detection process and we corrected them following the curation guidelines (Section~\ref{subsec:curation-guidelines}).
Then, we examined the corresponding training data corrected by the interface (Section~\ref{subsec:feedback-loop-training-data}) and obtained a set of 352 training data examples for our ML models. 
We call the obtained dataset \emph{curation} to be distinguished from the original SuperMat dataset which is referred to as \emph{base}.

We prepared our experiment using SciBERT~\cite{Beltagy2019SciBERT} that we fine-tuned for our downstream task as in~\cite{lfoppiano2023automatic}. 
We trained five models that we evaluated using a fixed holdout dataset from SuperMat averaging the results to smooth out the fluctuations. 
We use the DeLFT (Deep Learning For Text)~\cite{DeLFT} library for training, evaluating, and managing the models for prediction.  
A model can be trained with two different strategies: 
\begin{enumerate}
    \item \emph{``from scratch''}: when the model is initialised randomly. We denote this strategy with an \emph{(s)}.
    \item \emph{``incremental''}: when the initial model weights are taken from an already existing model. We denote this strategy with an \emph{(i)}.
\end{enumerate}
The latter can be seen as a way to ``continue'' the training from a specific checkpoint.
We thus define three different training protocols: 
\begin{enumerate}
    \item \textbf{base(s)}: using the \emph{base} dataset and training from scratch (s).
    \item \textbf{(base+curation)(s)}: using both the \emph{base} and \emph{curation} datasets and training from scratch (s).
    \item \textbf{base(s)+(base+curation)(i)}: Using the \emph{base} dataset to train from scratch (s), and then continuing the training with the \emph{curation} dataset (i).
\end{enumerate}
We merge ``curation'' with the base dataset because the curation dataset is very small compared to ``base'', and we want to avoid catastrophic forgetting~\cite{overcoming-kirkpatrick-etal-2016} or overfitting.

The trained models are then tested using a fixed holdout dataset that we designed in our previous work~\cite{lfoppiano2023automatic} and the evaluation scores are shown in Table~\ref{tab:evaluation-curation-training2}.

This experiment demonstrates that with only 352 examples (2\% of the SuperMat dataset) comprising 1846 additional entities (11\% of the entities from the SuperMat dataset) (Table~\ref{tab:training-support}), we obtain an improvement of F1-score from 76.67\%\footnote{In our previous work~\cite{lfoppiano2023automatic} we reported 77.03\% F1-score. 
There is a slight decrease in absolute scores between DeLFT 0.2.8 and DeLFT 0.3.0. 
One cause may be the use of different hyperparameters in version 0.3.0 such as batch size and learning rate.
However, the most probable cause could be the impact of using the Huggingface tokenizers library which is suffering from quality issues \url{https://github.com/kermitt2/delft/issues/150}.} to values between 77.44\% (+0.77) and 77.48\% (+0.81) for (base+curation)(s) and base(s)+(base+curation)(i), respectively. 


This experiment gives interesting insight relative to the positive impact on the way we select the training data. 
However, there are some limitations: the \emph{curation} dataset is small compared to the \emph{base} dataset. This issue could be verified by correcting all the available training data, repeating this experiment, and studying the interpolation between the size of the two datasets and the obtained evaluation scores. 
A second limitation is that the hyperparameters we chose for our model, in particular, the learning rate and batch size could be still better tuned to obtain better results with the second and third training protocols.

\subsection{Data quality}
\label{sec:interface-evaluation}
We conducted an experiment to evaluate the effectiveness and accuracy of data curation using two methods: a) the user interface (\textit{interface}), and b) the ``traditional'' manual approach consisting of reading PDF documents and populating an Excel file (\textit{PDF documents}).

We selected a dataset of 15 papers, which we assigned to three curators — a senior researcher (SD), a PhD student (PS), and a master's student (MS). 
Each curator received 10 papers: half to be corrected with the \textit{interface} and half with the \textit{PDF Document} method. 
Overall, each pair of curators had 5 papers in common which they had to process using opposite methods.
For instance, if curator A receives paper 1 to be corrected with the \textit{interface}, curator B, who receives the same paper 1, will correct it with the \textit{PDF document} method.
After curation, a fourth individual manually reviewed the curated content. The raw data is available in the Appendix~\ref{app:interface-evaluation-raw}.

We evaluated the curation considering a double perspective: time and correctness. 
Time was calculated as the accumulated minutes required using each method. 
Correctness was assessed using standard measures such as precision, recall, and the F1-score.
Precision measures the accuracy of the extracted information, while recall assesses the ability to capture all expected information. F1-Score is a harmonic means of precision and recall. 

\subsubsection{Discussion}
Overall, both methods required the same accumulated time: 185 minutes using the \textit{interface} and 184 minutes using the \textit{PDF Document} method.
When the experiment was carried out, not all the curators were familiar with the \textit{interface} method. Although they had access to the user documentation, they had to get acquainted with the user interface, thus the accumulated 185 minutes included such activities. 

We examined the quality of the extracted data and we observed an improvement of +5.55\% in precision and a substantial +46.69\% in recall when using the \textit{interface} as compared with the \textit{PDF Document} method (Table~\ref{tab:evaluation-interface-correction}). 
The F1-score improved by 39.35\%.

The disparity in experience significantly influenced the accuracy of curation, particularly in terms of high-level skills. Senior researchers consistently achieved an average F1-Score approximately 13\% higher than other curators (see Table~\ref{tab:accuracy-by-experience}). Furthermore, we observed a modest improvement between master's students and PhD students. These findings indicate also that for large-scale projects, employing master students instead of PhD students may be a more cost-effective choice. Thus, using only a few senior researchers for the second round of validation (Section~\ref{subsec:manual_correction}).

Finally, the collected data suggest that all three curators had overall more corrected results by using the interface as illustrated in Table~\ref{tab:accuracy-by-experience-method}. 

The results of this experiment confirmed that our curation interface and workflow significantly improved the quality of the extracted data, with an astonishing improvement in recall, thus preventing curators from overlooking important information.

\section{Code availability}
This work is available at \url{https://github.com/lfoppiano/supercon2}. The repository contains the code of the SuperCon\textsuperscript{2} interface, the curation workflow, and the ingestion processes for harvesting the SuperCon\textsuperscript{2} Database of materials and properties. The guidelines are accessible at \url{https://supercon2.readthedocs.io}.

\section{Conclusions}
We built a semi-automatic staging area, called SuperCon\textsuperscript{2}, to validate efficiently new experimental records automatically collected from superconductor research articles (SuperCon\textsuperscript{2} Database~\cite{lfoppiano2023automatic}) before they are ingested into the existing, manually-build database of superconductors, SuperCon~\cite{ishii2023structuring}.
The system provides a curation workflow and a user interface (SuperCon\textsuperscript{2} Interface) tailored to efficiently support domain experts in data correction and validation with fast context switching and an enhanced PDF viewer.
Under the hood, the workflow ran ``anomaly detection'' to automatically identify outliers and a ``training data collector'' based on human corrections, to efficiently accumulate training data to be feedback to the ML model. 

Compared with the traditional manual approach of reading PDF documents and extracting information in an Excel file, SuperCon\textsuperscript{2} significantly improves the curation quality by approximately 6\% and +47\% for precision and recall, respectively.
In future, this work can be expanded to support other materials science domains such as magnetic materials, spintronic and thermoelectric research and expanding the evaluation to a larger dataset. 



\section*{Acknowledgements}
Our warmest thanks to Patrice Lopez, the author of Grobid~\cite{GROBID}, DeLFT~\cite{DeLFT}, and other open-source projects for his continuous support and inspiration with ideas, suggestions, and fruitful discussions.
We thank Pedro Baptista de Castro for his support during this work. 
Special thanks to Erina Fujita for useful tips on the manuscript.

\section*{Funding}
This work was partly supported by MEXT Program: Data Creation and Utilization-Type Material Research and Development Project (Digital Transformation Initiative Center for Magnetic Materials) Grant Number JPMXP1122715503.

\section*{Notes on Contributors}
LF wrote the manuscript and KT helped with the editing. 
LF and POS discussed the ML results and experiments. 
LF implemented the workflow as a standalone service, and TM wrote the front end of the user interface. 
LF designed the user interface experiment with KT, TT and WS as curators.
KT led the materials-science work on the data with CS, TT and WS.
KT, TA, YT and MI revised the paper.
YT and MI supervised the work of the respective teams.

\bibliography{references}
\bibliographystyle{tfnlm}

\section*{Figures \& Tables}


\begin{table}[ht]
\centering\small
\caption{F1-score from the evaluation of the fine-tuned SciBERT models. The training is performed with three different approaches. 
The \emph{base} dataset is the original dataset described in~\cite{foppiano2021supermat}, and the \emph{curation} dataset is automatically collected based on the database corrections by the interface and manually corrected. \textit{s} indicate ``training from scratch'', while \textit{i} indicate ``incremental training''. 
The evaluation is performed using the same holdout dataset from SuperMat~\cite{foppiano2021supermat}. 
The results are averaged over 5 runs or train and evaluation. }
\begin{tabular}{lrrr}
\toprule
& \textbf{base(s)} & \textbf{(base+curation)(s)} & \textbf{base(s)+(base+curation)(i)} \\ 
\midrule
Nb total examples & 16902 & 17254 & 16902(s), 17254 (i)\\ 
\midrule
\texttt{<class>}        & 70.41         & \textbf{73.02}         & 71.86 \\ 
\texttt{<material>}     & 79.37         & 80.09         & \textbf{80.37} \\ 
\texttt{<me\_method>}   & 66.72         & 66.57         & \textbf{66.95} \\ 
\texttt{<pressure>}     & 46.43         & \textbf{48.42}         & 47.23 \\ 
\texttt{<tc>}           & 80.13         & \textbf{80.92}         & 80.34 \\ 
\texttt{<tcValue>}      & 78.29         & 78.41         & \textbf{79.73} \\ 
\midrule
\textbf{All (micro avg.)} & 76.67       & 77.44         & \textbf{77.48} \\ 
\midrule
\textbf{$\Delta$ avg. w/ baseline}& -   & +0.77     & \textbf{+0.81} \\ 
\bottomrule
\end{tabular}
\label{tab:evaluation-curation-training2}
\end{table}

\begin{table}[ht]
\centering
\small
\caption{Data support, the number of entities for each label in each of the datasets used for evaluating the ML models. The \emph{base} dataset is the original dataset described in~\cite{foppiano2021supermat}, and the \emph{curation} dataset is automatically collected based on the database corrections by the interface and manually corrected.}
\begin{tabular}{lccc}
\toprule
                        & \textbf{base}     & \textbf{base+curation}    & \textbf{$\Delta$}  \\ 
\midrule
\texttt{<class>}        & 1646              & 1732                      &  86                \\
\texttt{<material>}     & 6943              & 7580                      &  637               \\
\texttt{<me\_method>}   & 1883              & 1934                      &  51                \\
\texttt{<pressure>}     & 274               & 361                       &  87                \\
\texttt{<tc>}           & 3741              & 4269                      &  528               \\
\texttt{<tcValue>}      & 1099              & 1556                      &  457               \\
\midrule
\textbf{Total}          & 15586             & 17432                     & 1846               \\ 
\bottomrule
\end{tabular}
\label{tab:training-support}
\end{table}

\begin{table}[ht]
\centering\small
\caption{Evaluation scores (P: precision, R: recall, F1: F1-score) between the curation using the SuperCon\textsuperscript{2} interface (\textit{Interface}) and the traditional method of reading the PDF document (\textit{PDF document}). }
\begin{tabular}{lrrrr}
\toprule
    \textbf{Method}    & \textbf{P (\%)}   & \textbf{R (\%)}   & \textbf{F1 (\%)}  & \textbf{\# docs}   \\
    \midrule
    PDF document    & 87.83             & 45.61             & 52.67             & 15        \\
    Interface       & \textbf{93.38}    & \textbf{92.51}    & \textbf{92.02}    & 15        \\
    \bottomrule
\end{tabular}
\label{tab:evaluation-interface-correction}
\end{table}

\begin{table}[h]
\centering
\caption{Evaluation scores (P: precision, R: recall, F1: F1-score) aggregated by experience (MS: master student, PD: PhD student, SR: senior researcher). Each person corrected 10 documents.}
\begin{tabular}{lrrrrr}
\toprule
\textbf{Experience} & \textbf{P (\%)}   & \textbf{R (\%)}   & \textbf{F1 (\%)}  & \textbf{\#  docs} & \textbf{\# pages}\\
\midrule
MS      & 90.03             & 60.26             & 64.50           & 10  & 96    \\
PD      & 83.33             & 65.69             & 69.45           & 10  & 100   \\
SR      & \textbf{98.45}    & \textbf{81.22}    & \textbf{83.08}  & 10  & 96  \\
\bottomrule
\end{tabular}
\label{tab:accuracy-by-experience}
\end{table}

\begin{table}[h]
\centering\small
\caption{Evaluation scores (P: precision, R: recall, F1: F1-score) listed by experience (MS: master student, PD: PhD student, SR: senior researcher), and method (PDF document, Interface). }
\begin{tabular}{lcrrrrr}
\toprule
\textbf{Experience} & \textbf{Method} & \textbf{P (\%)} & \textbf{R (\%)} & 
\textbf{F1 (\%)}  & \textbf{\# docs} & \textbf{\# pages}\\
\midrule
\multirow{2}{*}{MS} & PDF Document & 94.58 & 36.55 & 48.67 & 6 & 46 \\
 & Interface & 83.19 & 95.83 & 88.25 & 4 & 50 \\
\midrule
\multirow{2}{*}{PD} & PDF Document & 70.00 & 48.51 & 50.78 & 5 & 49 \\
 & Interface & 96.67 & 82.86 & 88.11 & 5 & 51\\
\midrule
\multirow{2}{*}{SR} & PDF Document & \textbf{100.00} & 55.56 & 61.03 & 4 & 51\\
 & Interface & 97.42 & \textbf{98.33} & \textbf{97.78} & 6 & 45\\
\bottomrule
\end{tabular}
\label{tab:accuracy-by-experience-method}
\end{table}

\clearpage

\appendix

\section{Evaluation}
\label{app:interface-evaluation-raw}

\begin{table}[ht]
\centering
\small
\caption{Timetable recording the time spent for each of the 15 articles. Each row indicates the time and the event (Start, Finish) from each of the curators: Master Student (MD), PhD Student (PD), and Senior Researcher (SR). Duration is expressed in minutes.}
\label{tab:timetable-details}
\begin{tabular}{cc|cc|c}
Time & Event & Document ID & Curator & Duration (mins) \\
\toprule
14:40 & Start   & 02bf1b3db9 & PS  & 0  \\
14:49 & Finish  & 02bf1b3db9 & PS  & 9  \\
14:53 & Start   & 00b50fc0a8 & PS  & 0  \\
14:58 & Finish  & 00b50fc0a8 & PS  & 5  \\
14:37 & Start   & 0aa1b3161f & MS  & 0  \\
14:50 & Start   & 0454e07f64 & SR  & 0  \\
14:58 & Finish  & 0454e07f64 & SR  & 8  \\
15:01 & Start   & 02cbc58819 & PS  & 0  \\
15:06 & Start   & 00c32076f4 & SR  & 0  \\
15:07 & Finish  & 0aa1b3161f & MS  & 30 \\
15:08 & Finish  & 02cbc58819 & PS  & 7  \\
15:08 & Start   & 044939701d & PS  & 0  \\
15:12 & Start   & 0021fd339f & MS  & 0  \\
15:15 & Finish  & 00c32076f4 & SR  & 9  \\
15:17 & Finish  & 044939701d & PS  & 9  \\
15:17 & Start   & 08e1cb8f4f & PS  & 0  \\
15:20 & Start   & 0c7d3163ea & SR  & 0  \\
15:31 & Finish  & 08e1cb8f4f & PS  & 14 \\
15:32 & Finish  & 0021fd339f & MS  & 20 \\
15:32 & Start   & 039105663f & MS  & 0  \\
15:37 & Finish  & 0c7d3163ea & SR  & 17 \\
15:53 & Finish  & 039105663f & MS  & 21 \\
15:55 & Start   & 02c4f00127 & MS  & 0  \\
15:58 & Start   & 0454e07f64 & PS  & 0  \\
16:02 & Start   & 0da5febabf & SR  & 0  \\
16:08 & Finish  & 0454e07f64 & PS  & 10 \\
16:09 & Finish  & 02c4f00127 & MS  & 14 \\
16:11 & Finish  & 0da5febabf & SR  & 9  \\
16:11 & Start   & 0012333581 & SR  & 0  \\
16:12 & Start   & 00c32076f4 & PS  & 0  \\
16:18 & Start   & 021c413172 & MS  & 0  \\
16:22 & Finish  & 00c32076f4 & PS  & 10 \\
16:23 & Start   & 0c7d3163ea & PS  & 0  \\
16:30 & Finish  & 0012333581 & SR  & 19 \\
16:32 & Finish  & 021c413172 & MS  & 14 \\
16:37 & Start   & 02bf1b3db9 & MS  & 0  \\
16:38 & Finish  & 0c7d3163ea & PS  & 15 \\
17:32 & Finish  & 0021fd339f & SR  & 12 \\
17:34 & Start   & 039105663f & SR  & 0  \\
17:55 & Finish  & 039105663f & SR  & 21 \\
17:56 & Start   & 02c4f00127 & SR  & 0  \\
18:00 & Finish  & 02c4f00127 & SR  & 4  \\
18:00 & Start   & 021c413172 & SR  & 0  \\
18:09 & Finish  & 021c413172 & SR  & 9  \\
\end{tabular}
\end{table}

\begin{table}[ht]
\centering
\small
\caption{Evaluation scores obtained for each document and method (I: Interface, P: PDF) combination. TP: True positive, FP: False positive, FN: False negative. P: Precision, R: Recall, F1: F1-score }
\label{tab:curation-evaluation-detailed-results}
\begin{tabular}{cc|c|ccc|ccc}
\textbf{Document ID} & \# \textbf{pages} & \textbf{Method} & \# \textbf{TP}	& \# \textbf{FP}	& \# \textbf{FN}	& \textbf{P}	& \textbf{R}	& \textbf{F1} \\
\toprule
\multicolumn{9}{c}{Senior Researcher (SR)}\\
\midrule
0454e07f64  & 4     & I     & 6     & 0     &   0   & 100.00    & 100.00    & 100.00 \\
00c32076f4  & 13    & P     & 8     & 0     &   0   & 100.00    & 100.00    & 100.00 \\
0c7d3163ea  & 9     & I     & 13    & 1     &   0   & 92.86     & 100.00    & 96.30 \\
0da5febabf  & 11    & P     & 8     & 0     &   1   & 100.00    & 88.89     & 94.12 \\
0012333581  & 13    & I     & 11    & 0     &   0   & 100.00    & 100.00    & 100.00 \\
0aa1b3161f  & 5     & I     & 9     & 0     &   1   & 100.00    & 90.00     & 94.74 \\
0021fd339f  & 14    & P     & 4     & 0     &   8   & 100.00    & 33.33     & 50.00 \\
039105663f  & 9     & I     & 11    & 1     &   0   & 91.67     & 100.00    & 95.65 \\
02c4f00127  & 13    & P     & 0     & 0     &   3   & 100.00    & 0.00      & 0.00 \\
021c413172  & 5     & I     & 15    & 0     &   0   & 100.00    & 100.00    & 100.00 \\
\midrule
\multicolumn{9}{c}{PhD Student (PS)}\\
\midrule
02bf1b3db9  & 7     & I     & 5     & 0     &   2   & 100.00    & 71.43     & 83.33 \\
00b50fc0a8  & 11    & P     & 2     & 0     &   7   & 100.00    & 22.22     & 36.36 \\
02cbc58819  & 4     & I     & 4     & 0     &   3   & 100.00    & 57.14     & 72.73 \\
044939701d  & 12    & P     & 4     & 0     &   2   & 100.00    & 66.67     & 80.00 \\
08e1cb8f4f  & 16    & I     & 5     & 1     &   1   & 83.33     & 85.71     & 84.51 \\
0454e07f64  & 4     & P     & 0     & 1     &   5   & 0.00      & 16.67     & 0.00 \\
00c32076f4  & 13    & I     & 8     & 0     &   0   & 100.00    & 100.00    & 100.00 \\
0c7d3163ea  & 9     & P     & 9     & 0     &   5   & 100.00    & 64.29     & 78.26 \\
0da5febabf  & 11    & I     & 9     & 0     &   0   & 100.00    & 100.00    & 100.00 \\
0012333581  & 13    & P     & 4     & 4     &   3   & 50.00     & 72.73     & 59.26 \\
\midrule
\multicolumn{9}{c}{Master Student (MS)}\\
\midrule
0aa1b3161f  & 5     & P     & 1     & 0     &   9   & 100.00    & 10.00     & 18.18 \\
0021fd339f  & 14    & I     & 12    & 3     &   3   & 80.00     & 100.00    & 88.89 \\
039105663f  & 9     & P     & 4     & 1     &   7   & 80.00     & 41.67     & 54.79 \\
02c4f00127  & 13    & I     & 3     & 1     &   1   & 75.00     & 100.00    & 85.71 \\
021c413172  & 5     & P     & 7     & 1     &   7   & 87.50     & 53.33     & 66.27 \\
02bf1b3db9  & 7     & P     & 2     & 0     &   5   & 100.00    & 28.57     & 44.44 \\
00b50fc0a8  & 11    & I     & 7     & 2     &   0   & 77.78     & 100.00    & 87.50 \\
02cbc58819  & 4     & P     & 5     & 0     &   2   & 100.00    & 71.43     & 83.33 \\
044939701d  & 12    & I     & 5     & 0     &   1   & 100.00    & 83.33     & 90.91 \\
08e1cb8f4f  & 16    & P     & 1     & 0     &   6   & 100.00    & 14.29     & 25.00 \\
\bottomrule
\end{tabular}
\end{table}

\end{document}